\documentclass[conference]{IEEEtran}

\usepackage{graphicx} 
\usepackage{amsmath} 
\usepackage{mathtools}
\usepackage{cite}
\usepackage{lipsum}
\usepackage[font=small,skip=2pt]{caption}
\usepackage[caption=false]{subfig}

\begin{document}

\title{Pattern Generation for Walking on Slippery Terrains}

\author{\IEEEauthorblockN{Majid Khadiv\IEEEauthorrefmark{1}\IEEEauthorrefmark{2}\IEEEauthorrefmark{4},
S. Ali A. Moosavian\IEEEauthorrefmark{1},
Alexander Herzog\IEEEauthorrefmark{2}, and
Ludovic Righetti\IEEEauthorrefmark{2}\IEEEauthorrefmark{3}}
\IEEEauthorblockA{\IEEEauthorrefmark{1}Department of Mechanical Engineering, K. N. Toosi University of Technology, Tehran, Iran}
\IEEEauthorblockA{\IEEEauthorrefmark{2}Autonomous Motion Department, Max-Planck Institute for Intelligent Systems, Germany}
\IEEEauthorblockA{\IEEEauthorrefmark{3}Department of Mechanical and Aerospace Engineering, Department of Electrical and\\ Computer Engineering, New York University, USA}
\IEEEauthorblockA{\IEEEauthorrefmark{4}Corresponding author: mkhadiv@mail.kntu.ac.ir, majid.khadiv@tuebingen.mpg.de}}

\maketitle

\begin{abstract}
In this paper, we extend state of the art Model Predictive Control (MPC) approaches to generate safe bipedal walking on slippery surfaces. In this setting, we formulate walking as a trade off between realizing a desired walking velocity and preserving robust foot-ground contact. Exploiting this formulation inside MPC, we show that safe walking on various flat terrains can be achieved by compromising three main attributes, i. e. walking velocity tracking, the Zero Moment Point (ZMP) modulation, and the Required Coefficient of Friction (RCoF) regulation. Simulation results show that increasing the walking velocity increases the possibility of slippage, while reducing the slippage possibility conflicts with reducing the tip-over possibility of the contact and vice versa. 
\end{abstract}


\section{Introduction}

In order to take part in our daily life, humanoid robots should be able to perform various tasks reliably on challenging terrains. Since slip-related falls with very fast dynamics can cause severe damages, special care should be taken to prevent slippage during walking on surfaces with low available Coefficient of Friction (CoF). However, most walking planners based on abstract models for bipedal robots assume infinite available CoF and generate walking patterns without taking care of slippage avoidance constraints.

Trajectory optimization has become the prevailing approach for real-time gait generation of biped robots \cite{kajita2003biped,wieber2006trajectory,diedam2008online,herdt2010online,faraji2014robust,feng2016robust,khadiv2016step,khadiv2017robust}. Among various approaches, Model Predictive Control (MPC) has been shown to be a very effective tool. In this approach, the current measurement of states can be used to adapt gait variables to be robust against disturbances. It is shown that different walking pattern generation approaches based on analytical methods \cite{morisawa2007experimentation,englsberger2015three,khadiv2016stepping} and optimization techniques  \cite{herdt2010online,faraji2014robust,feng2016robust} can be seen as different variants of the same MPC problem \cite{wieber2016modeling}. 

Pioneered by the preview control of the Zero Moment Point (ZMP) \cite{kajita2003biped}, several modifications have been proposed to make this approach more versatile and robust \cite{wieber2006trajectory,diedam2008online,herdt2010online}. In these works and most of other approaches \cite{morisawa2007experimentation,faraji2014robust,feng2016robust,englsberger2015three,khadiv2016step,khadiv2017robust,khadiv2016stepping}, feasibility of the walking pattern is guaranteed by enforcing the ZMP to be inside the support polygon, while friction is assumed to be enough. Although this assumption usually results in feasible walking patterns on normal surfaces, it could be violated on challenging slippery terrains.

Studies carried out on human motion analysis show that subjects adapt their gait variables when they anticipate a reduction on the available CoF \cite{fong2009human,nagano2013biomechanical,chambers2013changes}. For bipedal walking, Kajita et al.\cite{kajita2004biped} showed that by increasing the weight of the Center of Mass (CoM) acceleration in the preview control scheme, they could decrease the Required Coefficient of Friction (RCoF). \cite{khadiv2015optimal} proposed an offline optimization procedure to generate walking patterns requiring low RCoF. \cite{li2015integral} suggested a simple adaptation of the generated CoM trajectory to avoid slippage of the stance foot. In \cite{brandao2016footstep}, the authors trained an infinite mixture of linear experts to fit a model between the gait variables as inputs, and the RCoF, the work done by the CoM and a function of constraints as outputs. They used a large number of simulations to train this model, and employed it to control walking on slippery terrains in a simulation environment.

As demonstrated by previous work, taking into account friction is important to ensure safe walking in various environments. However, to the best of our knowledge, state of the art walking pattern generators based on preview of the ZMP are not yet able to take this into account. In this paper, we propose an online walking pattern generator based on the MPC approach from \cite{herdt2010online} which explicitly takes into account friction constraints. In this setting, we formulate walking as a trade-off among walking velocity tracking, robustness to foot slippage, and robustness to foot tip-over. We consider feasibility constraints of the ZMP, RCoF, and step locations. Doing so, our gait planner automatically adapts the step locations and CoM trajectory for walking on various surfaces. 
We briefly present the MPC formulation from \cite{herdt2010online} in Section II, and extend the approach to friction constraints in  Section III. In Section IV, we present simulation experiments and discuss the properties of our algorithm. We conclude in Section V.

\section{Original MPC \cite{herdt2010online}}
In this section, we briefly outline the MPC approach proposed in \cite{herdt2010online} which is an extension of \cite{kajita2003biped,wieber2006trajectory,diedam2008online}. In this approach, a preview ($NT$) of piecewise constant CoM jerk over time intervals $(\dddot X_{k},\dddot Y_{k})$, as well as the future $m$ footstep locations $(X_{k}^f,Y_{k}^f)$  are considered as decision variables to minimize the CoM jerk as well as velocity tracking error and deviations from a desired ZMP:
\begin{align}
\label{Opt_orig}
J(u_k) =    &\frac{\alpha}{2}   \Vert \dddot X_{k}  \Vert^2  + \frac{\beta}{2}  \Vert \dot X_{k+1}-\dot X_{k+1}^{ref } \Vert^2+ \frac{\gamma}{2}  \Vert  Z_{k+1}^x- Z_{k+1}^{x_{ref} } \Vert^2  \nonumber\\
 &\frac{\alpha}{2}   \Vert \dddot Y_{k}  \Vert^2  + \frac{\beta}{2}  \Vert \dot Y_{k+1}-\dot Y_{k+1}^{ref } \Vert^2+ \frac{\gamma}{2}  \Vert  Z_{k+1}^y- Z_{k+1}^{y_{ref }} \Vert^2   
\end{align}

where:
\begin{align}
\label{des_var}
u_k=\begin{pmatrix} \dddot X_{k} \\  X_{k}^f \\ \dddot Y_{k}  \\ X_{k}^f   	\end{pmatrix}    \;, \; 
\dddot X_{k}=\begin{pmatrix} \dddot x_{k} \\  \vdots \\ \dddot x_{k+N-1}     	\end{pmatrix}  \; , \; 
 X_{k}^f=\begin{pmatrix}  x_{k}^f \\  \vdots \\  x_{k+m-1}^f     	\end{pmatrix}    \nonumber\\
X_{k+1}=\begin{pmatrix} x_{k+1} \\  \vdots \\  x_{k+N}     	\end{pmatrix}     ,  
\dot X_{k+1}=\begin{pmatrix} \dot x_{k+1} \\  \vdots \\ \dot x_{k+N}     	\end{pmatrix}     , 
 Z_{k+1}^x=\begin{pmatrix}  z_{k+1}^x \\  \vdots \\  z_{k+N}^x     	\end{pmatrix}
\end{align}

In these equations, $Z$ encodes the ZMP preview and the superscript $ref$ stands for the reference trajectory. $\alpha, \beta$ and $\gamma$ are positive numbers. $(x_k,y_k)$ is the CoM position in $k$th time interval, and $(x_k^f,y_k^f)$ represents the next footstep location in the series of $m$ future steps. For the lateral direction $Y$, the same formulation as (\ref{des_var}) can be obtained. Using (\ref{des_var}), the position and velocity of the CoM as well as the ZMP position on the preview horizon can be written as a function of the current state of the CoM $(\hat x_k , \hat y_k)$ and the jerk vector:
\begin{align}
\label{matrices}
X_{k+1}&=P_{ps} \hat x_k+P_{pu} \dddot X_k \nonumber\\
\dot X_{k+1}&=P_{vs} \hat x_k+P_{vu} \dddot X_k \\
Z_{k+1}&=P_{zs} \hat x_k+P_{zu} \dddot X_k \nonumber
\end{align}

where the matrices $P_{ps},P_{vs},P_{zs} \in \Re^{N \times 3}$, and  $P_{pu},P_{vu},P_{zu} \in \Re^{N \times N}$ represent discrete integration over time and can be computed recursively \cite{herdt2010online}. Furthermore, the ZMP at each time interval $(z_k^x,z_k^y)$ is computed using the Linear Inverted Pendulum Model (LIPM) dynamics \cite{kajita20013d}:
\begin{align}
\label{zmp_eq}
z_k^x=\begin{pmatrix} 1 & 0 & -h/g  	\end{pmatrix}   \hat x_k \nonumber\\
z_k^y=\begin{pmatrix} 1 & 0 & -h/g  	\end{pmatrix}   \hat y_k
\end{align}

In this equation, $h$ specifies the pendulum height and $g$ is the gravity constant. The reference ZMP trajectory in (\ref{Opt_orig}) can be formulated as a function of the current stance foot location $(x_k^{fc},y_k^{fc})$ and the following foot locations in the horizon $(X_k^f ,Y_k^f )$:
\begin{align}
\label{zmp_foot}
Z_{k+1}^{x_{ref}}&=U_{k+1}^c x_k^{fc}+U_{k+1} X_k^f  \nonumber\\
Z_{k+1}^{y_{ref}}&=U_{k+1}^c y_k^{fc}+U_{k+1} Y_k^f  
\end{align}

where $U_{k+1}^c \in \Re^{N}$ and $U_{k+1} \in \Re^{N \times m}$ correspond respectively to the current and future step locations ($m$ steps) \cite{herdt2010online}.
Finally, the problem of finding CoM jerk and footstep locations can be canonically written as a Quadratic Program (QP) as follows:
\begin{align}
\label{can_orig}
\underset{u_k}{\text{min.}}   \quad    \frac{1}{2}    u_k^T Q_k u_k+p_k^T u_k
\end{align}

where
\begin{align}
\label{hess_orig}
Q_k=\begin{pmatrix}   Q_k^\prime & 0 \\ 0 &  Q_k^\prime  \end{pmatrix}   \nonumber
\end{align}

\begin{align}
Q_k^\prime=\begin{pmatrix}
\alpha I+\beta P_{vu}^T P_{vu} +\gamma P_{zu}^T P_{zu} & -\gamma P_{zu}^T U_{k+1} \\
 -\gamma U_{k+1}^T P_{zu}  &  \gamma U_{k+1}^TU_{k+1}
\end{pmatrix} 
\end{align}

and
\begin{align}
\label{grad_orig}
p_k=
\begin{pmatrix}
\beta P_{vu}^T (P_{vs} \hat x-\dot X_{k+1}^{ref}) +\gamma P_{zu}^T (P_{zs} \hat x-U_{k+1}^c X_k^{fc}) \\
 -\gamma U_{k+1}^T (P_{zs} \hat x-U_{k+1}^c X_k^{fc})\\ 
\beta P_{vu}^T (P_{vs} \hat y-\dot Y_{k+1}^{ref}) +\gamma P_{zu}^T (P_{zs} \hat y-U_{k+1}^c Y_k^{fc}) \\
 -\gamma U_{k+1}^T (P_{zs} \hat y-U_{k+1}^c Y_k^{fc})
\end{pmatrix} 
\end{align}

This QP is solved under linear inequality constraints on the ZMP and step locations \cite{herdt2010online} to automatically generate the CoM trajectory and footstep locations in real-time.

\section{Proposed formulation}

The method presented in the last section does not take into account frictional constraints. As a result, it is possible that the generated walking pattern becomes unfeasible, because the robot would start slipping. In this section, we generalize this approach by taking into account friction. First, we use the LIPM equations and relate the RCoF as a linear function of the gait variables. Then, we add to the cost function a term that minimizes friction forces. Finally, we impose the friction cone constraints to guarantee slippage avoidance for walking on various terrains.

\subsection{RCoF}

The LIPM linearizes the biped dynamics by assuming fixed CoM height and zero angular momentum rate around the CoM \cite{kajita20013d}. We used (\ref{zmp_eq}) to compute the ZMP in terms of the CoM position and acceleration. Using the LIPM dynamics, we can also relate the horizontal and vertical forces in terms of the CoM and ZMP positions at time interval $k$:
\begin{align}
\label{friction}
f_k^x=\frac{x_k-z_k^x}{h}f_k^z \nonumber\\
f_k^y=\frac{y_k-z_k^y}{h}f_k^z
\end{align}

\begin{figure}
\centering
\includegraphics[clip,trim=5cm 17cm 8cm 4cm,width=6cm]{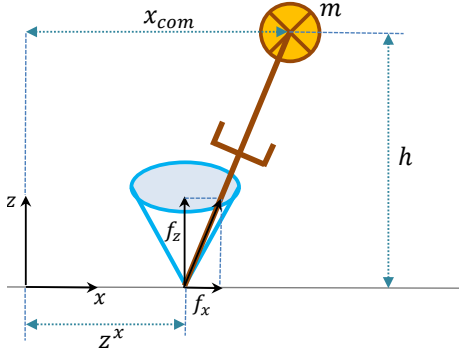}
\caption{The LIPM and corresponding friction cone}
\vspace{-1.5em}
\label{LIPM}
\end{figure}

This equation is obtained by taking the moment about the CoM (Fig. \ref{LIPM}). We define $\mu_k^{x_{req}}$ and $\mu_k^{y_{req}}$ as the RCoF in the sagittal and lateral directions:
\begin{align}
\label{forces}
\mu_k^{x_{req}} \coloneqq \frac{f_k^x}{f_k^z}=\frac{x_k-z_k^x}{h} \nonumber\\
\mu_k^{y_{req}} \coloneqq \frac{f_k^y}{f_k^z}=\frac{y_k-z_k^y}{h}
\end{align}

This equation yields interesting information on the relation between the RCoF and the ZMP. It implies that to reduce the RCoF for a walking pattern, the distance between the CoM and the ZMP should be reduced. The distance between the CoM and the ZMP can be interpreted as a measure of how dynamic the motion is. As a result, if the walking pattern is static and the CoM and the ZMP coincide during motion, no friction is needed to realize walking. However, the more dynamic the motion is, the more RCoF is needed to realize the motion. As a result for walking with high velocity and large steps  where motions are more dynamic, the distance between the CoM and the ZMP increases and as a result the RCoF is increased. It should be noted that these analyses are valid for the LIPM abstraction of a biped, and change of the  CoM height and rate of angular momentum around the CoM affect the RCoF as well.

For a walking pattern to be feasible, the friction cone constraint should hold:
\begin{align}
\label{mu_quad}
(\mu_k^{x_{req}})^2+(\mu_k^{y_{req}})^2 \leq\mu_k^{av}
\end{align}

where $\mu_k^{av}$ is the available CoF between the feet and the ground surface. When the robot traverses surfaces with different slipperiness characteristics,  $\mu_k^{av}$ changes and the walking pattern should be adapted to realize a feasible motion.

\begin{figure}
\centering
\includegraphics[clip,trim=8cm 18.7cm 7.3cm 4cm,width=5cm]{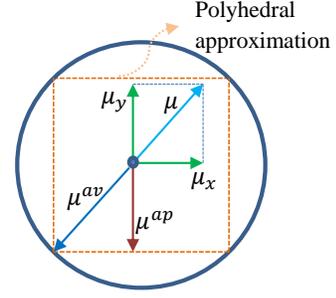}
\caption{Polyhedral approximation of frictional constraints}
\vspace{-1.5em}
\label{Friction_cone}
\end{figure}

To keep the MPC structure linear, we use a conservative polyhedral approximation ($\mu_k^{ap}$) of (\ref{mu_quad}) which boils down to two sets of independent linear constraints (Fig. \ref{Friction_cone}):
\begin{align}
\label{mu_lin}
|\mu_k^{x_{req}}| \leq \mu_k^{ap}=\frac{\sqrt2}{2} \mu_k^{av} \nonumber\\
|\mu_k^{y_{req}}| \leq \mu_k^{ap}=\frac{\sqrt2}{2} \mu_k^{av}
\end{align}

\subsection{Proposed cost function}

We define a cost function which is a trade-off among three main features, i. e. velocity tracking, tip-over avoidance, and slippage avoidance. Our main goal is to generate walking patterns consistent with the desired velocity, while minimizing the possibility of contact instability. Hence, the proposed cost function is:
\begin{align}
\label{Opt_prop}
J_{f} =   & \frac{\beta}{2}  \Vert \dot X_{k+1}-\dot X_{k+1}^{ref } \Vert^2+ \frac{\gamma}{2}  \Vert  Z_{k+1}^x- Z_{k+1}^{x_{ref} } \Vert^2 + \frac{\delta}{2} \Vert M_{k+1}^{x_{req}} \Vert^2  \nonumber\\
 &\frac{\beta}{2}  \Vert \dot Y_{k+1}-\dot Y_{k+1}^{ref } \Vert^2+ \frac{\gamma}{2}  \Vert  Z_{k+1}^y- Z_{k+1}^{y_{ref }} \Vert^2   + \frac{\delta}{2} \Vert M_{k+1}^{y_{req}} \Vert^2
\end{align}

where $\delta$ is the weight of the RCoF. Moreover, $M_{k+1}^{req}$ is the RCoF vector in the receding horizon and can be computed using (\ref{forces}) for sagittal and lateral directions:
\begin{align}
\label{RCoF_hor}
 M_{k+1}^{x_{req}}=\begin{pmatrix} \mu_{k+1}^{x_{req}} \\  \vdots \\ \mu_{k+N}^{x_{req}}  \end{pmatrix}   =  \frac{1}{h}\begin{pmatrix} x_{k+1}-z_{k+1}^x \\  \vdots \\  x_{k+N}-z_{k+N}^x \end{pmatrix} \nonumber\\
 M_{k+1}^{y_{req}}=\begin{pmatrix} \mu_{k+1}^{y_{req}} \\  \vdots \\ \mu_{k+N}^{y_{req}}  \end{pmatrix}   =  \frac{1}{h}\begin{pmatrix} y_{k+1}-z_{k+1}^y \\  \vdots \\  y_{k+N}-z_{k+N}^y \end{pmatrix}
\end{align}

The first term in the cost function of  (\ref{Opt_prop}) enforces the CoM velocity to be as close as possible to the desired walking velocity.  The second term tries to keep the ZMP close to the center of the stance foot and can be interpreted as a tip-over stability margin. The third term keeps the resultant interaction force away from boundaries of the friction cone. The last three terms have the same meaning for the lateral direction. It seems necessary to note that as it is mentioned in \cite{wieber2016modeling}, minimizing any derivative of the CoM position can be considered as a sufficient condition of viablity. Hence, the velocity tracking terms in the cost function not only enforce the velocity tracking task, but also guarantee viability of the motion.

The weights in (\ref{Opt_prop}) are chosen depending on the terrain that the robot is traversing. For instance, walking on slippery terrains needs maximum robustness against slippage. As a result, $\delta$ should be increased to alleviate the possibility of slip-related falls. This may be achieved at the cost of degrading the walking velocity tracking and increasing the risk of foot tip-over. On the other hand, for the cases that the robot should traverse surfaces with partial footholds and uncertain supporting area, $\gamma$ should be increased. Hence, safe walking on such surfaces is obtained at the cost of more velocity tracking error (decreasing the walking velocity and increasing the fluctuations on the CoM velocity), and less slippage robustness. As a result, compromising these three factors adapts the generated motion consistent with the situation, while tuning the weights can be seen as an interesting learning problem.

Using (\ref{matrices}) and (\ref{RCoF_hor}), the vector $M_{k+1}^{req}$ can be written down in matrix form in terms of current state of the CoM and the future jerks:
\begin{align}
\label{RCoF}
M_{k+1}^{x_{req}}&=\frac{1}{h}((P_{ps}-P_{zs}) \hat x_k+(P_{pu}-P_{zu}) \dddot X_k)\nonumber\\
M_{k+1}^{y_{req}}&=\frac{1}{h}((P_{ps}-P_{zs}) \hat y_k+(P_{pu}-P_{zu}) \dddot Y_k)
\end{align}

As a result the Hessian in (\ref{hess_orig}) and gradient in (\ref{grad_orig}) can be modified as:
\begin{align}
\label{hess_grad_prop}
Q_k^\prime(1,1)&=\beta P_{vu}^T P_{vu} +\gamma P_{zu}^T P_{zu}+\delta(P_{pu}-P_{zu})^T(P_{pu}-P_{zu}) \nonumber\\
p_k(1)&=\beta P_{vu}^T (P_{vs} \hat x-\dot X_{k+1}^{ref}) +\gamma P_{zu}^T (P_{zs} \hat x-U_{k+1}^c X_k^{fc}) \nonumber\\&+\delta(P_{pu}-P_{zu})^T(P_{ps}-P_{zs}) \hat x\nonumber\\
p_k(3)&=\beta P_{vu}^T (P_{vs} \hat y-\dot Y_{k+1}^{ref}) +\gamma P_{zu}^T (P_{zs} \hat y-U_{k+1}^c Y_k^{fc}) \nonumber\\&+\delta(P_{pu}-P_{zu})^T(P_{ps}-P_{zs}) \hat y
\end{align}

It should be noted that we eliminated $1/h^2$ which is a constant and multiplies to the weight $\delta$ in the Hessian and gradient.

\begin{figure}
\centering

\subfloat[Desired and actual walking velocity]{%
  \includegraphics[clip,trim=2.7cm 5.3cm 3.2cm 1cm,width=9cm]{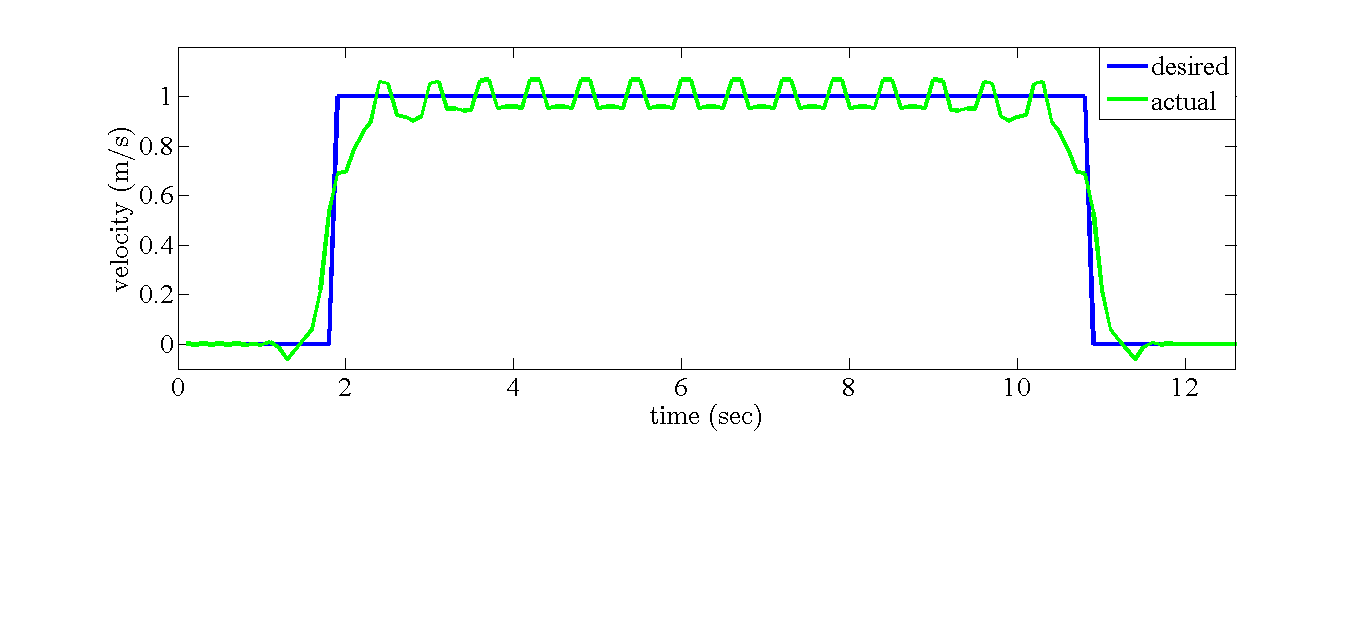}
}

\subfloat[Required Coefficient of Friction (RCoF)]{%
  \includegraphics[clip,trim=2.7cm 4cm 3.2cm 1cm,width=9cm]{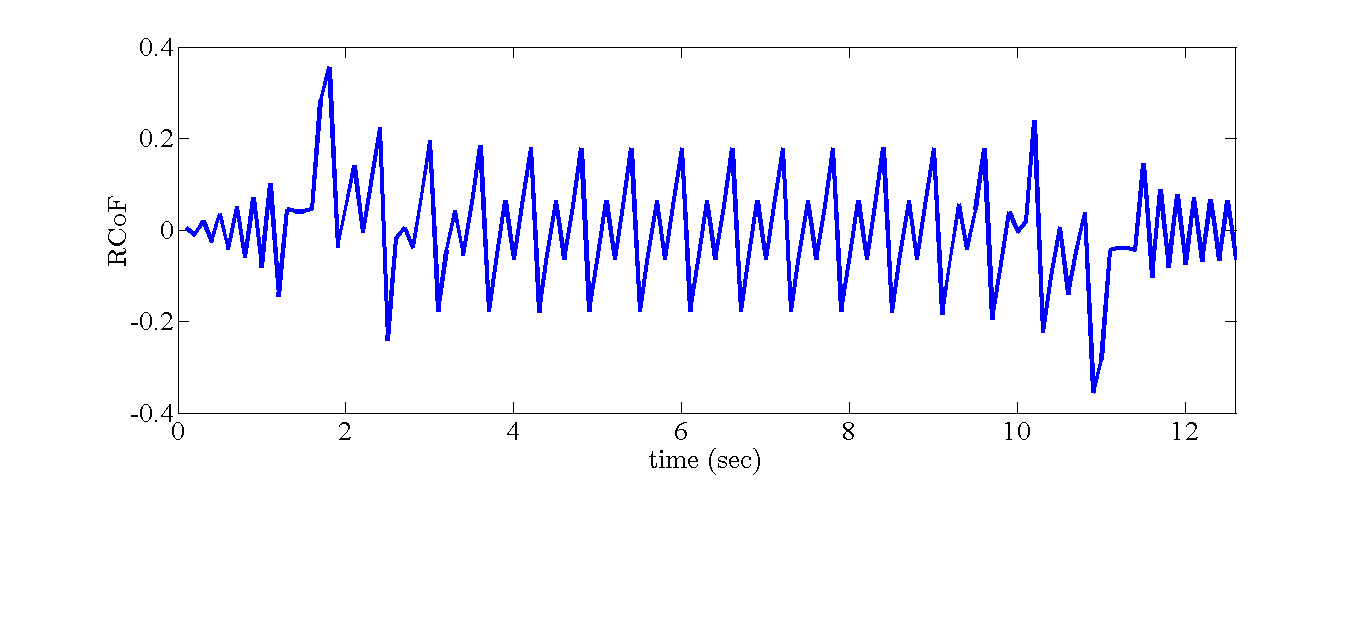}
}

\subfloat[Footstep locations and ZMP trajectory]{%
  \includegraphics[clip,trim=0cm 7.8cm .5cm 1.5cm,width=9cm]{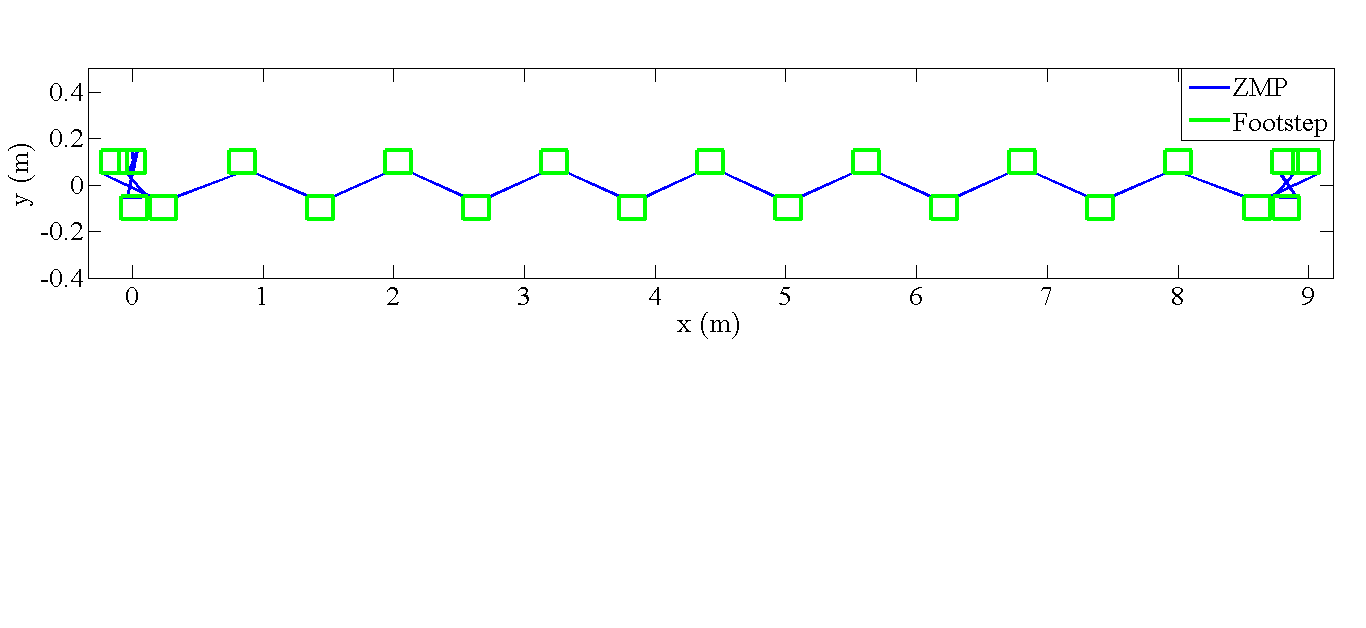}
}
\caption{The first simulation scenario with $\beta=1$ and $\gamma=\delta=0$.}
\vspace{-1em}
\label{just_beta}
\end{figure}

\subsection{Frictional constraints}

We use the pyramid approximation of the friction cone (\ref{mu_lin}) to construct linear constraints which allows us to solve a QP to generate the motion. These constraints can be formulated as:
\begin{align}
\label{fric_cons}
\begin{pmatrix}   P_{pu}-P_{zu} & 0_{N\times m} & 0_{N\times N}& 0_{N\times m} \\ 0_{N\times N}& 0_{N\times m} &  P_{pu}-P_{zu} & 0_{N\times m}  \end{pmatrix}& u_k \leq   \nonumber\\
\begin{pmatrix}    M^{ap} h -(P_{ps}-P_{zs})\hat x \\  M^{ap} h -(P_{ps}-P_{zs})\hat y\end{pmatrix}
\end{align}

and
\begin{align}
\label{fric_cons2}
\begin{pmatrix}   P_{zu}-P_{pu} & 0_{N\times m} & 0_{N\times N}& 0_{N\times m} \\ 0_{N\times N}& 0_{N\times m} &  P_{zu}-P_{pu} & 0_{N\times m}  \end{pmatrix}& u_k \leq   \nonumber\\
\begin{pmatrix}    M^{ap} h + (P_{ps}-P_{zs})\hat x \\  M^{ap} h + (P_{ps}-P_{zs})\hat y\end{pmatrix}
\end{align}

where $M^{ap} \in \Re^{N}$ is the vector of polyhedral approximation of the available friction coefficient in the horizon (see Fig. \ref{Friction_cone}).

\begin{figure}
\centering

\subfloat[Desired and actual walking velocity]{%
  \includegraphics[clip,trim=2.7cm 5.3cm 3.2cm 1cm,width=9cm]{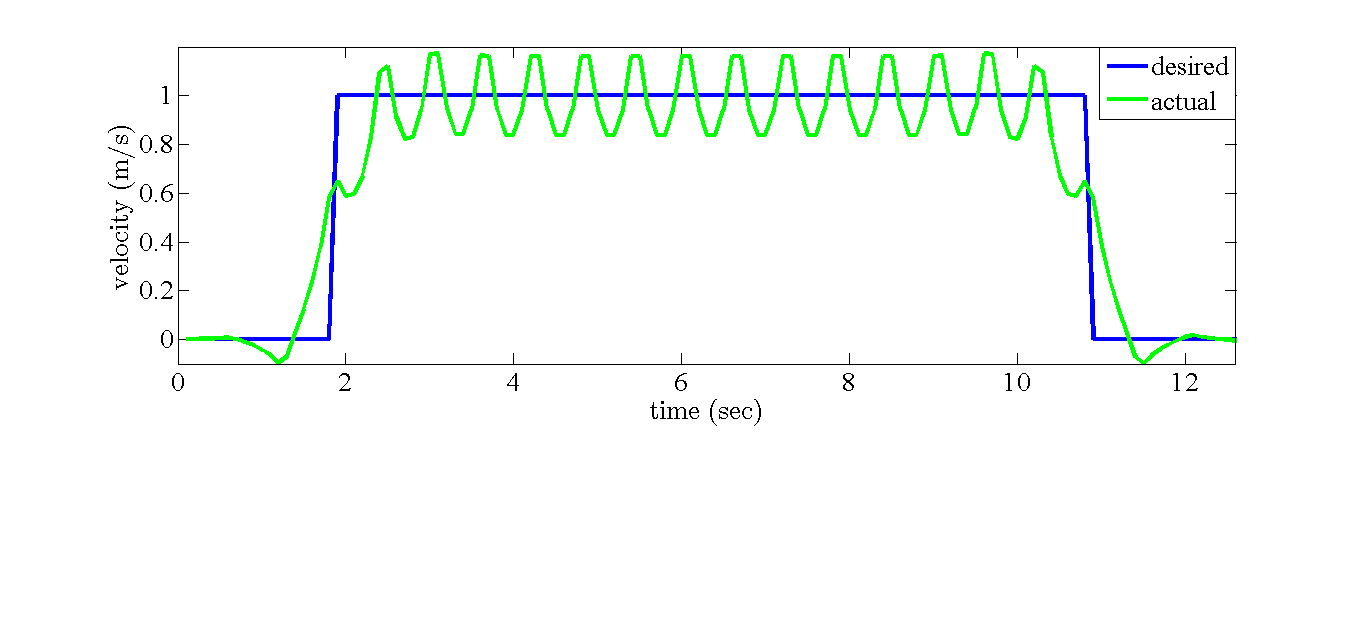}
}

\subfloat[Required Coefficient of Friction (RCoF)]{%
  \includegraphics[clip,trim=2.7cm 4cm 3.2cm 1cm,width=9cm]{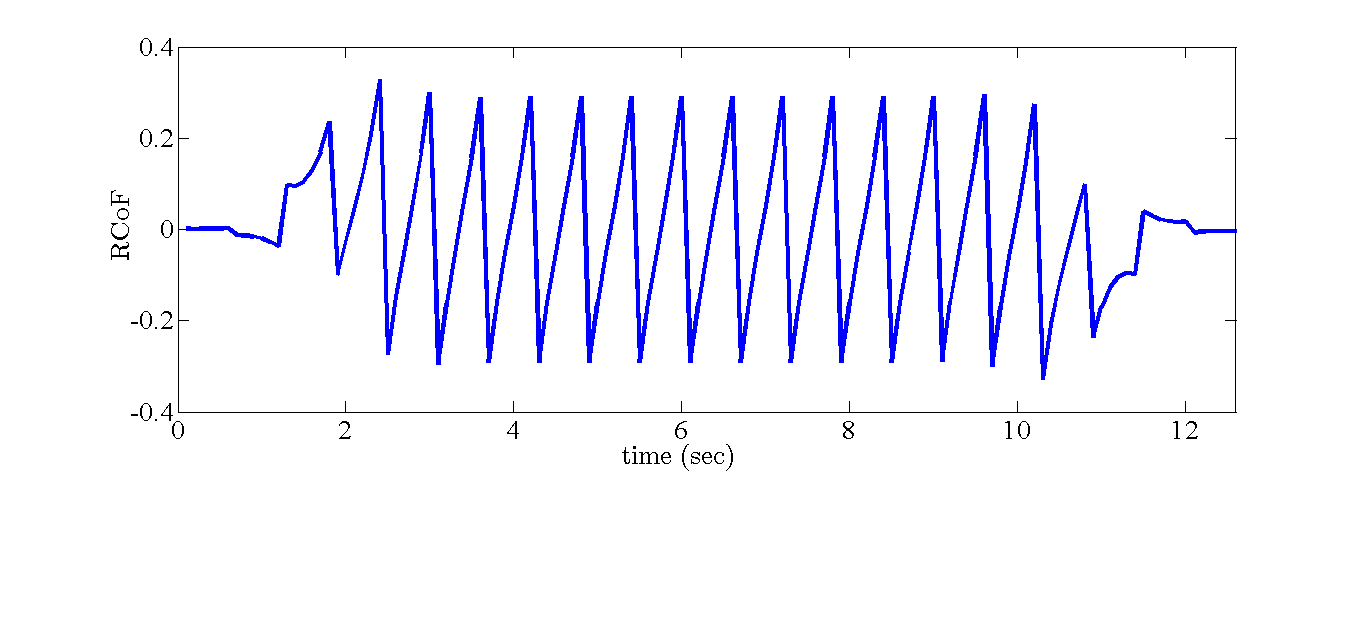}
}

\subfloat[Footstep locations and ZMP trajectory]{%
  \includegraphics[clip,trim=0cm 8.3cm 0.5cm 1cm,width=9cm]{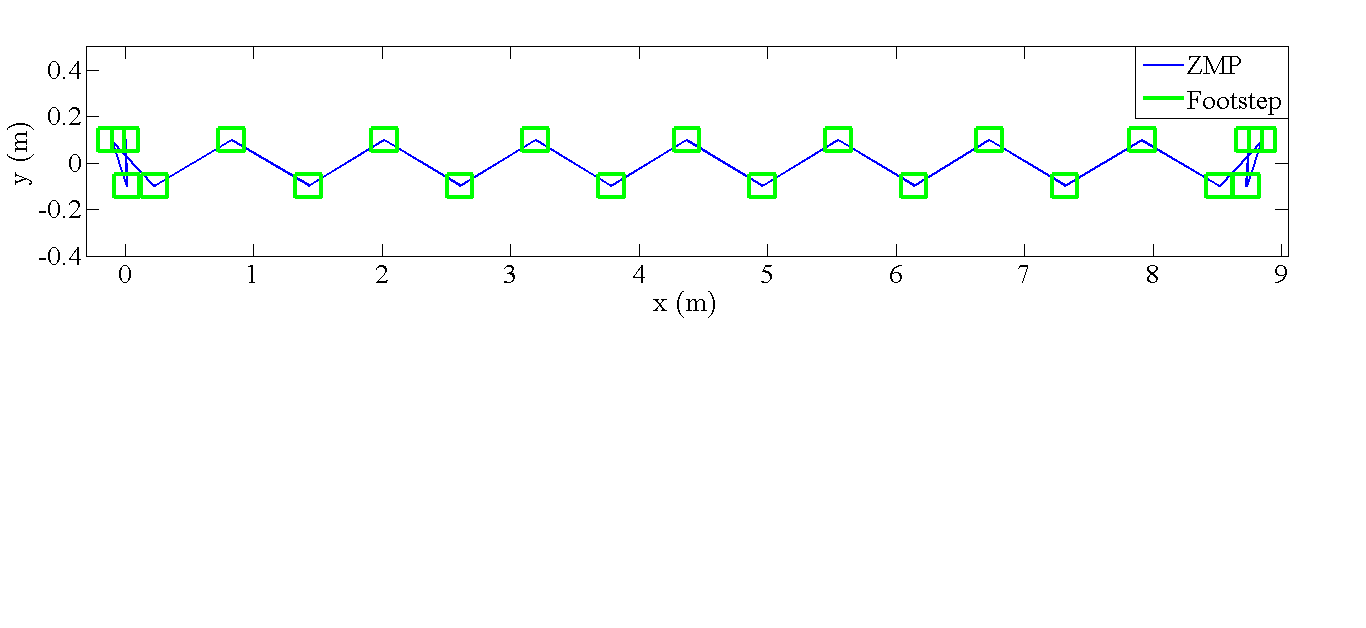}
}
\caption{The first simulation scenario with $\beta=1$, $\gamma=100$ and $\delta=0$. }
\vspace{-1em}
\label{beta_gamma}
\end{figure}

\section{Simulation results}
In this section, we present two simulation scenarios based on the LIPM abstraction of a biped robot. The physical properties of the model and the robot constraints as well as the gait parameters are given in TABLE \ref{physical_properties}. The preview horizon of the MPC for all simulations is two steps.

In the first simulation scenario, we demonstrate walking with different weights of the cost function in (\ref{Opt_prop}). In this scenario, the robot starts stepping in place, then a desired forward walking velocity command $v_{des}=1 m/s$ is given. Finally, by commanding zero walking velocity, the robot resumes stepping in place again. 

\begin{table}[h]
\caption{Physical properties of the model and the gait parameters}
\label{physical_properties}
\vspace{-1em}
\begin{center}
\begin{tabular}{|c|c|c|}
\hline
\it parameter & \it description& \it value\\
\hline
$T$ & Length of time interval& 0.1 ($sec$)\\
\hline
$h$ & LIPM height& 80 ($cm$)\\
\hline
$T_{ss}$ & single support duration& 0.5 ($sec$)\\
\hline
$T_{ds}$ & double support duration& 0.1 ($sec$)\\
\hline
$a$ & Foot length& 20 ($cm$)\\
\hline
$b$ & Foot width& 10 ($cm$)\\
\hline
$L_{max}$ & maximum step length& 60 ($cm$)\\
\hline
$W_{max}$ & maximum step width& 40 ($cm$)\\
\hline
\end{tabular}
\end{center}
\end{table}

We start with just considering the velocity tracking in the cost function, which means $\beta=1, \gamma=\delta=0$. In this case, hard constraints on the ZMP and RCoF guarantee the motion feasibility. The available CoF is considered to be $\mu^{av}=0.7$ ($\mu^{ap}=0.5$) and the foot length and width for the ZMP constraints are given in TABLE \ref{physical_properties}. A summary of illustrations for this case is presented in Fig. \ref{just_beta}. As it can be observed in Fig. \ref{just_beta} (a), the walking velocity is tracked well. Since the ZMP and RCoF are not considered in the cost function in this case, the optimizer does not trade off slippage and tip-over robustness, and just keeps the RCoF less than $\mu^{ap}=0.5$. That is why we can see significant changes in the RCoF in Fig. \ref{just_beta} (b), especially when the desired walking velocity is changed. Furthermore, as it can be seen in Fig. \ref{just_beta} (c), the ZMP moves most of the time on the boundaries of the support polygon which substantially increases the possibility of the stance foot tip-over. In fact, moving the ZMP on the edges of the support polygon decreases the walking velocity tracking error, but increases the possibility of fall due to foot slippage and tip-over.

\begin{figure}
\centering

\subfloat[Desired and actual walking velocity]{%
  \includegraphics[clip,trim=.8cm 0cm 1.2cm 1cm,width=8cm]{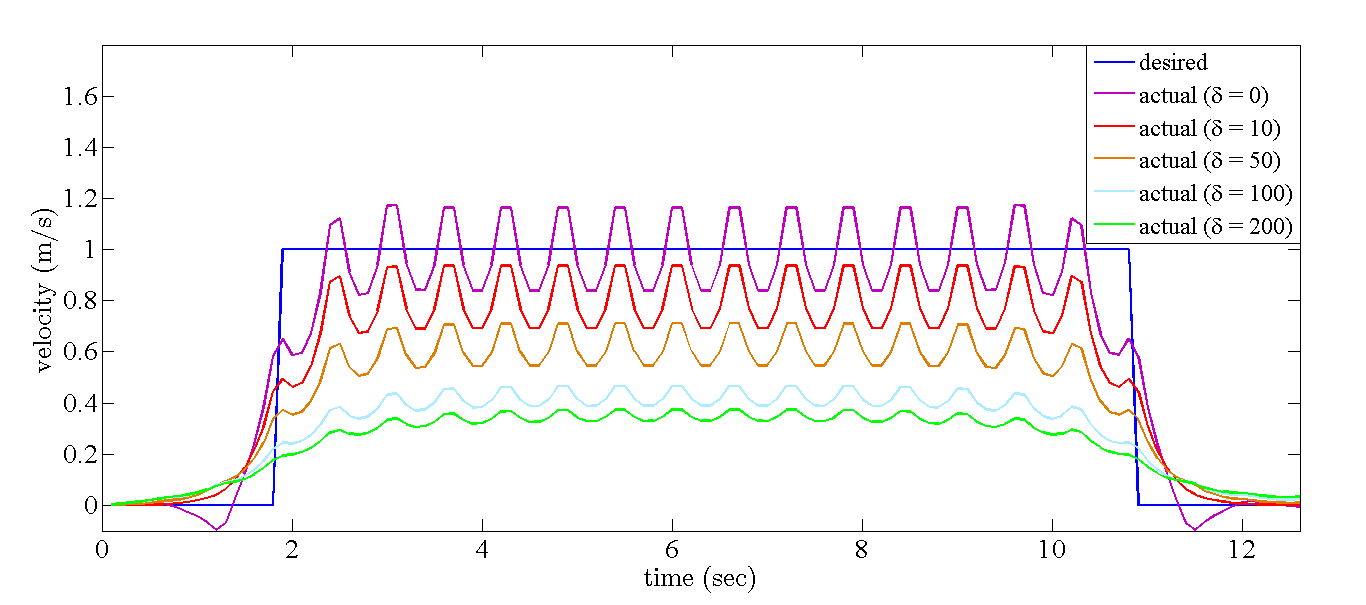}
}

\subfloat[Required Coefficient of Friction (RCoF)]{%
  \includegraphics[clip,trim=2.7cm 1.5cm 3cm .9cm,width=8cm]{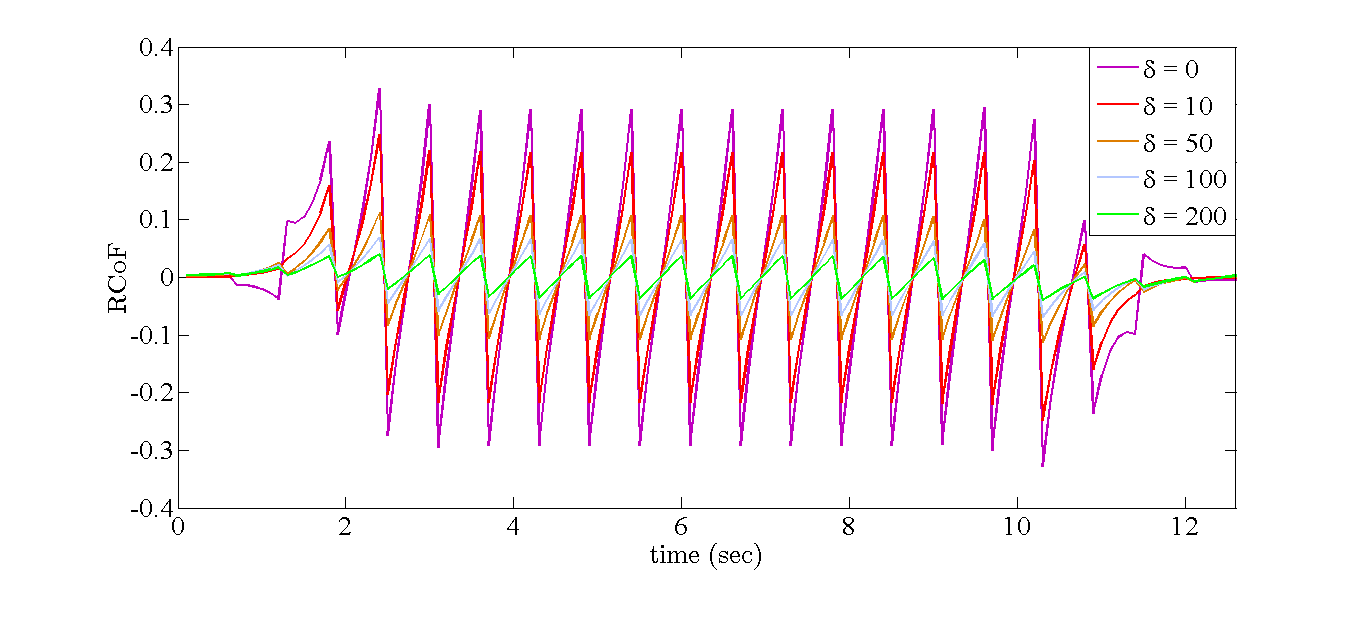}
}

\caption{The first simulation scenario with $\beta=1$, $\gamma=100$ and $\delta$ changes from zero to 200 }
\vspace{-1em}
\label{vel_mu_delta}
\end{figure}

Since falling down prevention is more important than desired walking velocity tracking, we need to make the walking patterns more robust against contact instability. We start with increasing the weight $\gamma$ to bring the ZMP closer to the center of the support polygon and reduce the possibility of foot tip-over. In Fig. \ref{beta_gamma}, the results for the case $\gamma=100$ are shown. As we can see in Fig. \ref{beta_gamma} (a), although the fluctuations on the actual velocity is increased compare to Fig. \ref{just_beta} (a), the mean walking velocity tracks the desired velocity well. Since in this case we give higher weight to the ZMP tracking compared to the walking velocity tracking, the actual walking velocity in this case is smoother which causes the RCoF to be smoother as well. We cannot see drastic changes in the RCoF when the desired walking velocity changes, however the maximum RCoF during steady walking with $v=1m/s$ is increased from about 0.2 to 0.35. As a result, enforcing the ZMP trajectory to be close to the center of the support polygon during walking (Fig. \ref{beta_gamma} (c)) is achieved at the cost of a negligible increase in the instantaneous walking velocity tracking error, but a considerable increase in the RCoF during steady forward walking. This experiment illustrates how minimizing deviations of the ZMP leads to higher friction forces.

\begin{figure}
\centering
  \includegraphics[clip,trim=0cm 4.3cm 0cm 1cm,width=9cm]{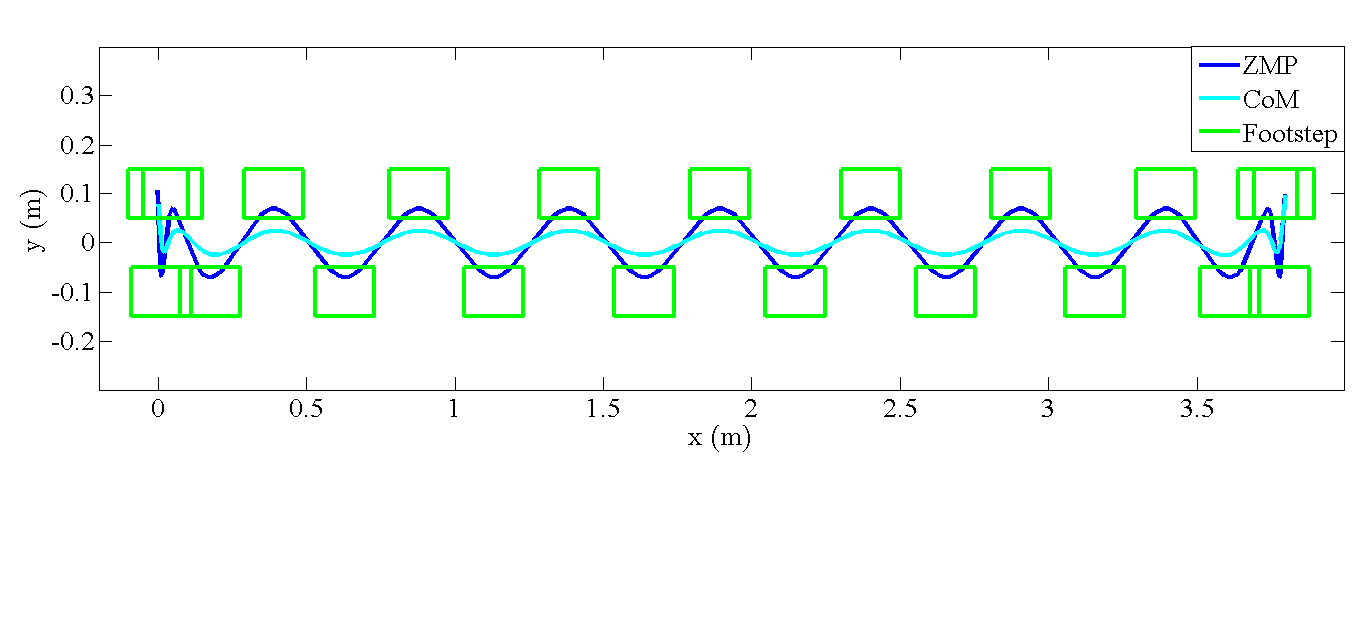}
\caption{Foot locations, CoM and ZMP trajectory for $\beta=1$, $\gamma=100$ and $\delta=100$}
\vspace{-1.5em}
\label{foot_com_zmp}
\end{figure}

In order to explore the effects of adding the RCoF term in the cost function (\ref{Opt_prop}), we start to increase its weight $\delta$ while keeping the other weights fixed. Fig. \ref{vel_mu_delta} shows the obtained results for changing $\delta$ from 0 to 200. We can see in Fig.  \ref{vel_mu_delta} (a) that by increasing the RCoF weight, the quality of walking velocity tracking is degraded considerably. This observation suggests that in order to increase the robustness against slippage, the walking velocity (step length for constant step duration) should be alleviated. In other words, increasing the walking velocity needs a more aggressive gait with more possibility of slippage. We can see in  Fig. \ref{vel_mu_delta} (b) that for decreasing the RCoF from 0.35 to less than 0.1, we need to decrease the walking velocity from $1 m/s$ to around $0.4 m/s$. Besides, as it can be deduced from (\ref{forces}), not only the velocity tracking is degraded, but the ZMP needs also to approach the CoM to decrease the RCoF (see Fig. \ref{foot_com_zmp}). This means that we can only decrease the RCoF at the cost of either degrading walking velocity tracking or reducing tip-over robustness, or potentially a combination of both. 

The analyses we carried out up to now suggest that as long as the velocity term exists in the cost function and the viability is guaranteed, we can find cost function weights that significantly change the gait properties. For instance, if the surface that the robot tends to traverse seems slippery based on the visual feedback, the robot could increase the weight $\delta$ to increase the robustness against slippage. On the other hand, in the situation where the footholds are limited or uncertain, $\gamma$ should be increased. Since changing these weights does not affect the viability of the motion, changing the weights can be done without any concern about falling. As a result, the approach allows to explore different surfaces properties. This could be useful, for example, to adapt the weights based on online measurements of the surface properties.

\begin{figure}
\centering

\subfloat[Desired and actual walking velocity]{%
  \includegraphics[clip,trim=2.7cm 3.6cm 3.2cm 1cm,width=8.9cm]{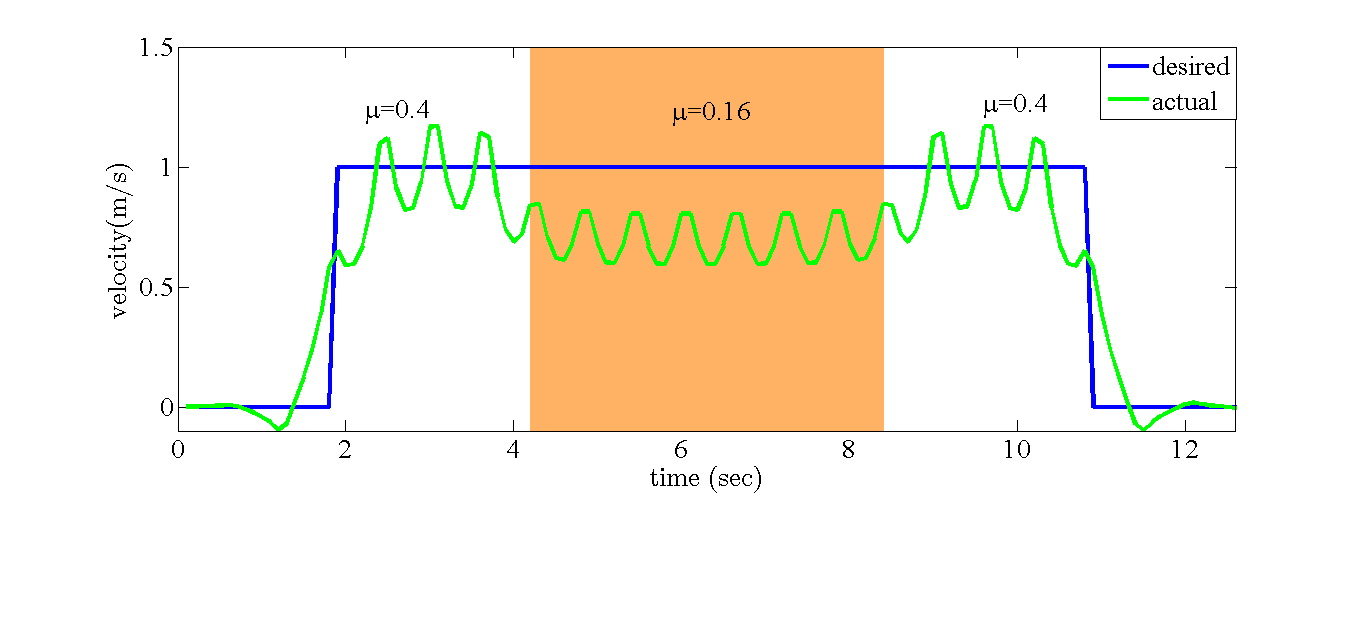}
}

\subfloat[Required Coefficient of Friction (RCoF)]{%
  \includegraphics[clip,trim=2.7cm 3.6cm 3.2cm 1cm,width=8.9cm]{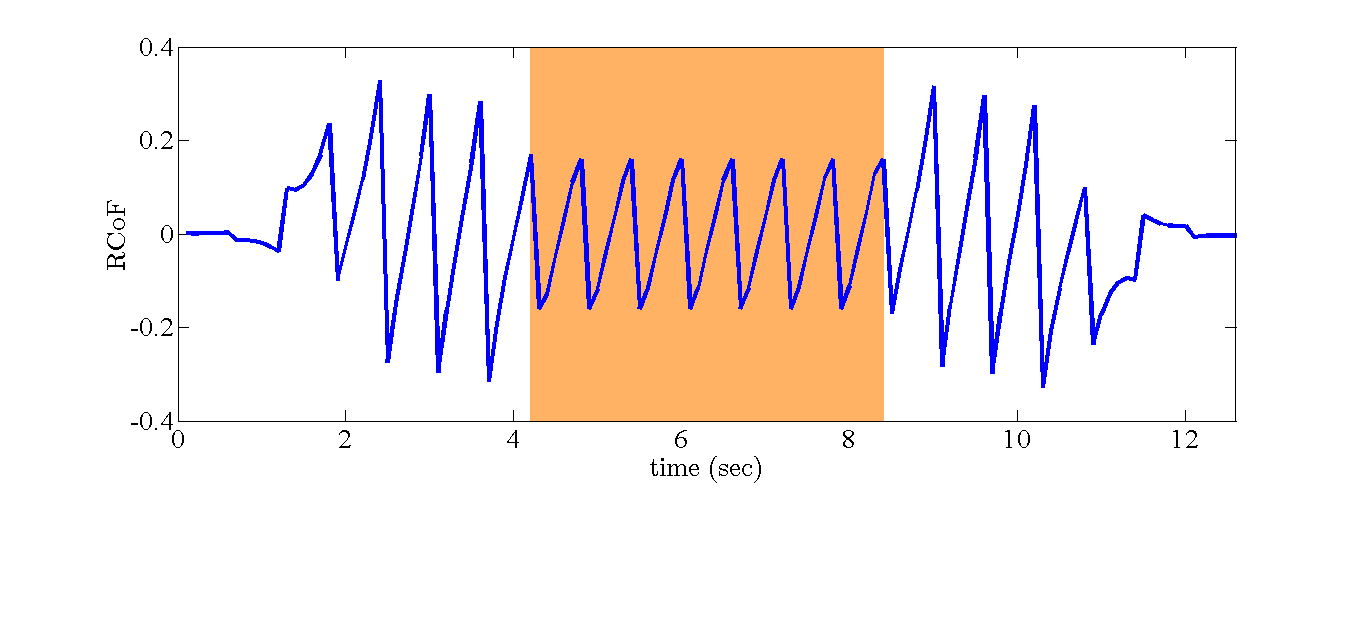}
}

\subfloat[Footstep locations and ZMP trajectory]{%
  \includegraphics[clip,trim=0.5cm 7.3cm 0.7cm 1cm,width=8.9cm]{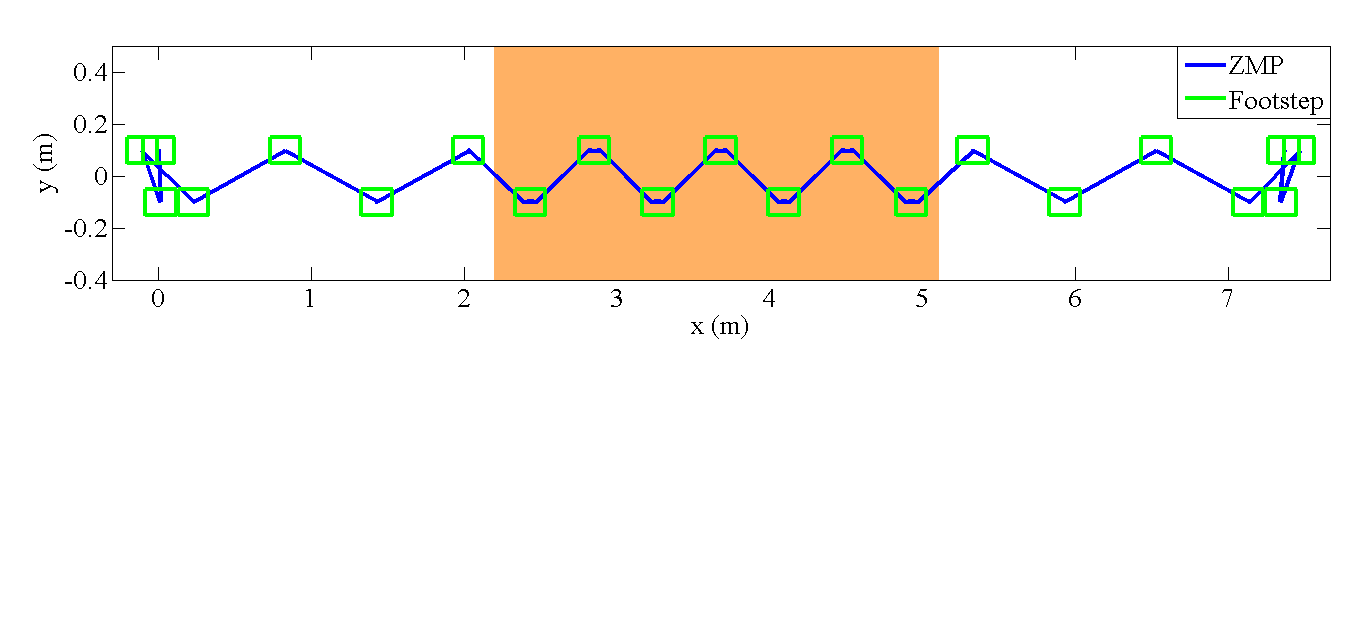}
}
\caption{Second simulation scenario ($\beta=1$, $\gamma=100$ and $\delta=1$): The robot starts stepping on a terrain with $\mu^{ap}=0.4$. After seven steps, the robot enters a slippery surface, while $\mu^{ap}=0.16$ and is known in advance. The optimizer automatically adapts the gait variables to bring the RCoF below 0.16 to make the walking pattern feasible.}
\vspace{-1em}
\label{mu_change}
\end{figure}

In the second scenario, we investigate a case where the surface CoF is known in advance. In this scenario, by setting $M^{ap}$ (consistent with $\mu^{av}$) appropriately in the preview horizon, we can ask the optimizer to automatically adapt the gait variables to traverse slippery surfaces. We assume that the robot starts stepping on a surface with $\mu^{av}=0.56$ ($\mu^{ap}=0.4$) and after taking seven steps, it suddenly enters a surface with $\mu^{av}=0.23$ ($\mu^{ap}=0.16$). The cost function weights in this scenario are $\beta=1$, $\gamma=100$ and $\delta=1$. As it is shown in Fig. \ref{mu_change}, once the robot enters the slippery terrain, the actual walking velocity is reduced (Fig \ref{mu_change} (a)) to bring the RCoF below 0.16 (Fig \ref{mu_change} (b)). This is achieved by decreasing the step length and letting the ZMP move from heel to toe such that the distance between the ZMP and CoM is reduced (Fig \ref{mu_change} (c)).  

\section{CONCLUSION}

In this paper, we proposed a walking pattern generator based on MPC which takes into account foot slippage avoidance. In this setting, walking is realized by compromising walking velocity tracking against contact stability robustness in the cost function, while friction cone constraints are applied using polyhedral approximation of the friction cone. By taking into account the frictional constraints, the walking pattern is adapted for the case when the robot enters a terrain with low available CoF. Simulation results show that walking velocity tracking conflicts with robustness against contact instability. In other words, to achieve walking with high velocities, we need to make the walking gaits more aggressive by increasing the RCoF and letting the ZMP approach the boundaries of the support polygon. Furthermore, decreasing the RCoF to gain more robustness against slippage is obtained at the cost of decreasing the robustness against foot tip-over. 

 \section*{Acknowledgment}
The third and fourth authors are supported by the Max-Planck Society, MPI-ETH center for learning systems and the European Research Council under the European Union's Horizon 2020 research and innovation program (grant agreement No 637935).

\bibliography{Master}
\bibliographystyle{IEEEtr}

\end{document}